\documentclass[conference]{IEEEtran}

\makeatletter
\def\ps@IEEEtitlepagestyle{%
  \def\@oddfoot{\mycopyrightnotice}%
  \def\@evenfoot{}%
}
\def\mycopyrightnotice{%
  {\footnotesize 979-8-3195-4351-6/26/\$31.00~\copyright~2026 IEEE\hfill}% <--- Change here
  \gdef\mycopyrightnotice{}
}

\usepackage{blindtext}
\usepackage{eso-pic}
\IEEEoverridecommandlockouts
\usepackage{cite}
\usepackage{amsmath,amssymb,amsfonts}
\usepackage{algorithmic}
\usepackage{graphicx}
\usepackage{textcomp}
\usepackage{xcolor}
\def\BibTeX{{\rm B\kern-.05em{\sc i\kern-.025em b}\kern-.08em
    T\kern-.1667em\lower.7ex\hbox{E}\kern-.125emX}}
    
\usepackage{eso-pic}
\newcommand\AtPageUpperMyright[1]{\AtPageUpperLeft{%
 \put(\LenToUnit{0.17\paperwidth},\LenToUnit{-2cm}){%
     \parbox{0.9\textwidth}{\raggedleft\fontsize{8}{11}\selectfont #1}}%
 }}%
\newcommand{\conf}[1]{%
\AddToShipoutPictureBG*{%
\AtPageUpperMyright{#1}
}
}

\begin{document}
\title{\vspace*{1cm} Multi-Terrain Mastery: A Comprehensive Controller for Bipedal Locomotion\\
\thanks{Funding for this work was in part provided by NSF Award No.~2118818 with the generous support of Jessy W Girzzle. The work of O. Dosunmu-Ogunbi was partially supported by a Rackham Merit Fellowship.}
}

\author{\IEEEauthorblockN{1\textsuperscript{st} Oluwami Dosunmu-Ogunbi}
\IEEEauthorblockA{\textit{Mechanical Engineering Department} \\
\textit{Ohio Northern University}\\
Ada, United States of America \\
0000-0002-9585-8119}
\and
\IEEEauthorblockN{2\textsuperscript{nd} Aayushi Shrivastava}
\IEEEauthorblockA{\textit{Mechanical Engineering Department} \\
\textit{University of California, Berkeley}\\
Berkeley, United States of America \\
aayushis@berkeley.edu}
% \and
% \and
% \and
% \and
}

\maketitle
\conf{\textit{  Proc. of the International Conference on Electrical, Computer, Communications and Mechatronics Engineering (ICECCME 2026) \\ 
15-17 October 2026, Bali, Indonesia}}

% Abstract
\begin{abstract}
% This document is a model and instructions for \LaTeX.
% This and the IEEEtran.cls file define the components of your paper [title, text, heads, etc.]. *CRITICAL: Do Not Use Symbols, Special Characters, Footnotes, 
% or Math in Paper Title or Abstract.

Advancing bipedal robots to navigate diverse terrains remains a significant challenge in robotics. Traditional locomotion controllers excel on specific surfaces but struggle across varied environments, limiting their practical applications. Given the unpredictable nature of real-world environments, a single controller capable of handling multiple terrains is ideal, eliminating the need for multiple specialized controllers. We propose a multi-terrain controller to enhance the versatility and robustness of bipedal locomotion. Building on previous work with a stance ankle motor for stability on inclined and rough surfaces, this paper extends capabilities to steep wet uneven grassy slopes, and compliant terrains such as sand, gravel, rocks, and constrained terrains like staircases. To address the unique demands of these terrains, we introduce a new impact map that is essential for maintaining performance and robustness against unseen terrains. We also discuss in detail the control structure for real-time deployment on the robot. We validate our controller on the 20 degree-of-freedom Cassie bipedal robot.
\end{abstract}

\begin{IEEEkeywords}
robotics
\end{IEEEkeywords}

% Sections

% Introduction/Motivation
\section{Introduction} 
\label{sec:introduction}

The development of robust and adaptable bipedal robots capable of navigating diverse terrains has been a long-standing challenge in robotics. Traditional locomotion controllers often excel on specific surfaces but struggle to maintain stability and efficiency across varied environments such as stairs, sand, gravel, rocks and grassy slopes. This limitation hinders the deployment of bipedal robots in real-world applications where environmental conditions are unpredictable and dynamic. To address this challenge, we propose a comprehensive multi-terrain controller designed to enhance the versatility and robustness of bipedal robot locomotion without utilizing any visual feedback.

% Furthermore, while our previous conference papers were constrained by page limits and thus focused on specific aspects of the controller, this comprehensive journal paper offers a holistic explanation of the entire control structure. 
% It details our implementation of UDP and the CasADi open-source tool for nonlinear optimization and differentiation, crucial for meeting the execution time requirements to deploy our controller on the Cassie bipedal robot. To ensure that we present our controller holistically, we necessarily reiterate information from \cite{dosunmu2023stair} and \cite{dosunmu2024demonstrating} as well as incorporate previously neglected information.

 \begin{figure}
    \centering
    \includegraphics[width=.45\textwidth]{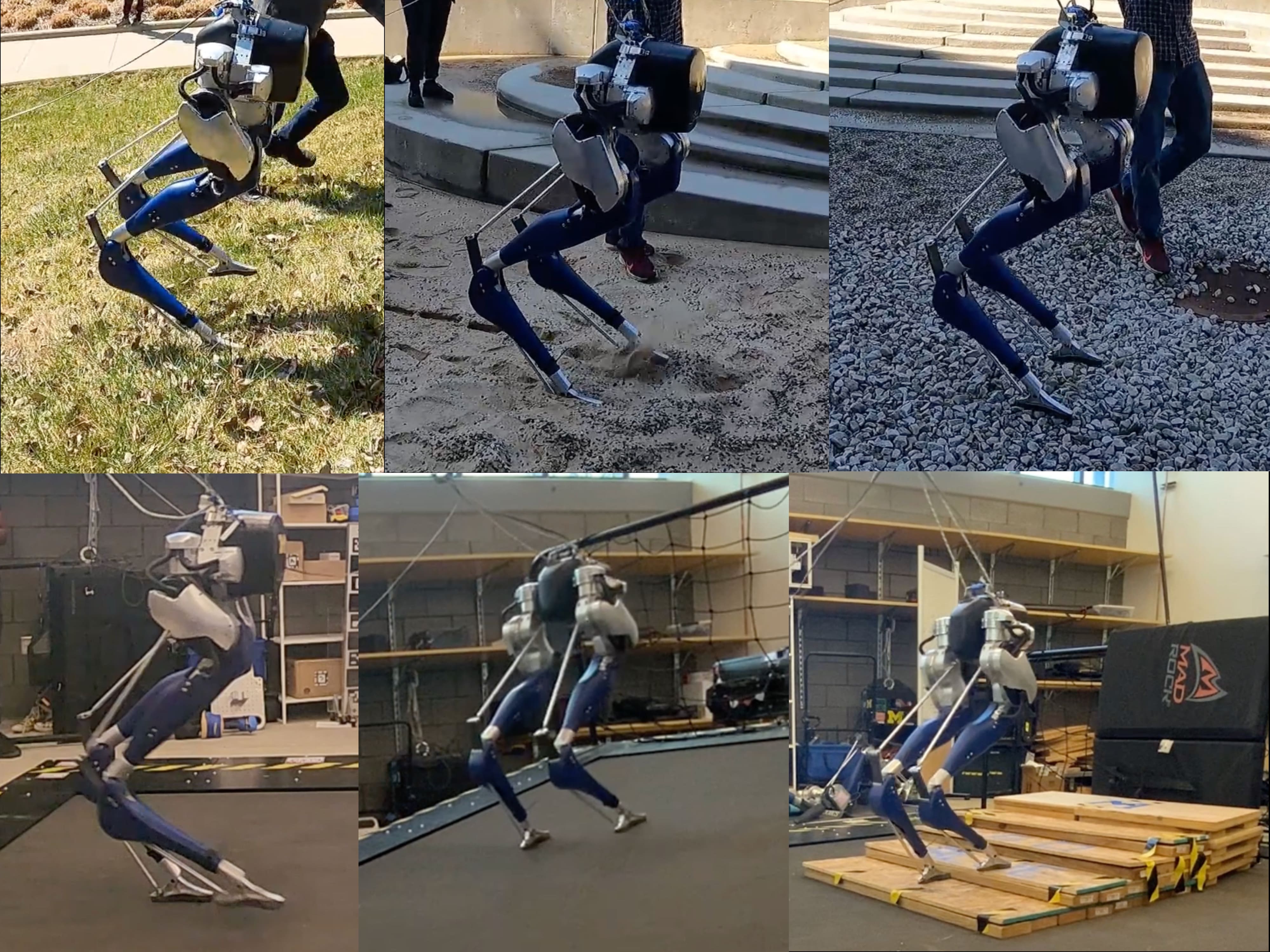}
    \caption{A collage of screenshots where the Cassie bipedal robot walks on grassy slope, sand, rocks,  moving walkway, inclined treadmill and stairs.}
    \label{fig:collage}
\end{figure}

% \subsection{Background}
% \label{sec:background}

The DARPA Robotic Challenge brought forth a new generation of bipedal walking controllers to real hardware systems. In \cite{zheng2013humanoid} the authors address the challenge of walking on rough surfaces such as grass, sand, and rock by using different gait patterns (a ``step-over" gait and ``ski-type" gait) as a global approach to maintain stability while compliant motion is used to solve the robust foot-holding problem on the fully-actuated Hubo 2+ robot. In \cite{hopkins2015design} the authors developed a controller to improve stability on dirt and stairs by modifying swing foot position based on the divergent component of motion and by modifying center of pressure trajectory for the fully-actuated ESCHER robot. 

A new generation of bipedal walking controllers soon followed in the steps of the DARPA era. In \cite{gong2019feedback}, the authors successfully test their angular momentum-based controller on the Cassie robot on grass, sand, snow, and burning bush environment. While the results are impressive, the controller does not utilize the stance ankle motor which makes it miss out on enhanced stability and robustness as is shown in this paper. In \cite{gora2023energy}, the authors developed an energy-based footstep planning controller for a bipedal robot on uneven deformable terrain using the nonlinear inverted pendulum model. In \cite{muenprasitivej2024bipedal}, the authors developed a novel hierarchical planning framework for bipedal navigation in rough and uncertain terrain environments using Gaussian process learning of uncertainty. The controller was successfully tested on a simulation model of the under-actuated Digit robot. The authors of \cite{Stephane2020Capturability} used variable-height inverted pendulum (VHIP) and show how it enables 3-D walking over uneven terrains based on capture inputs on a simulation model of the under-actuated humanoid robot. In \cite{Xiong2021}, the authors developed a novel controller to enable navigation over rough and challenging terrains for the actuated Spring Loaded Inverted Pendulum Model. In \cite{acosta2023ankle}, the authors have constrained footholds along with ankle torque to navigate different terrains. However, the controller's performance was reduced in outdoor environments due to the unreliability of perception in such settings. In \cite{Duan2024learning} the authors have shown sim-to-real learning for vision-based bipedal locomotion over challenging indoor terrains. Despite their strengths, these approaches tend to be terrain-specific and do not generalize effectively to all environments.

Previous works \cite{dosunmu2023stair}, \cite{dosunmu2024demonstrating}, utilize the often overlooked stance ankle motor to enhance walking stability on inclined and rough surfaces, as well as ascending stairs. These works demonstrated the modified Angular Momentum based Linear Inverted Pendulum (ALIP) model controller and its ability to navigate stairs in the SimMechanics simulation environment \cite{dosunmu2023stair} and traverse non-flat, non-stationary terrains \cite{dosunmu2024demonstrating}. In this paper, we extend the capabilities of the controller based on above mentioned works to operate on compliant terrains, such as sand, gravel, rocks, and steep, wet grassy slopes outdoors, in addition to stairs as shown in Figure \ref{fig:collage}. We introduce a new impact map essential for tackling these complex tasks without compromising the controller's existing functionalities. We offload the controller to a separate laptop to optimize the execution time of our controller. 

This paper demonstrates a multi-use, robust controller that not only functions on flat, inclined, and non-stationary environments \cite{dosunmu2024demonstrating}, but also stairs and compliant terrain such as sand, gravel, rocks, and an inclined wet, uneven, grassy slope.

\subsection{Contributions}
\label{sec:contributions}
The contributions of this paper are as follows:
\begin{enumerate}

\item Holistic description of the entire controller structure for bipedal robot walking on compliant and constrained terrain,
\item Derivation of new impact map based on linearizing over a nominal trajectory to improve controller robustness, 
\item Demonstration of improved controller over sand, gravel, rocks, and steep, wet grassy slope and stairs on the Cassie biped robot hardware.

\end{enumerate}

% Modeling the Cassie Bipedal Robot
% \input{Sections/ModelingCassie/modelingCassie}

% Controller Overview
\section{Development of the Controller}
\label{sec:controllerDevelopment}
In this section, we delve into our control philosophy, explore the concept of nominal periodic orbits, examine the principles behind our passivity-based and model predictive controllers, and detail our strategy for stabilizing lateral foot placement.

\begin{figure}
    \centering
    \includegraphics[width=0.5\textwidth]{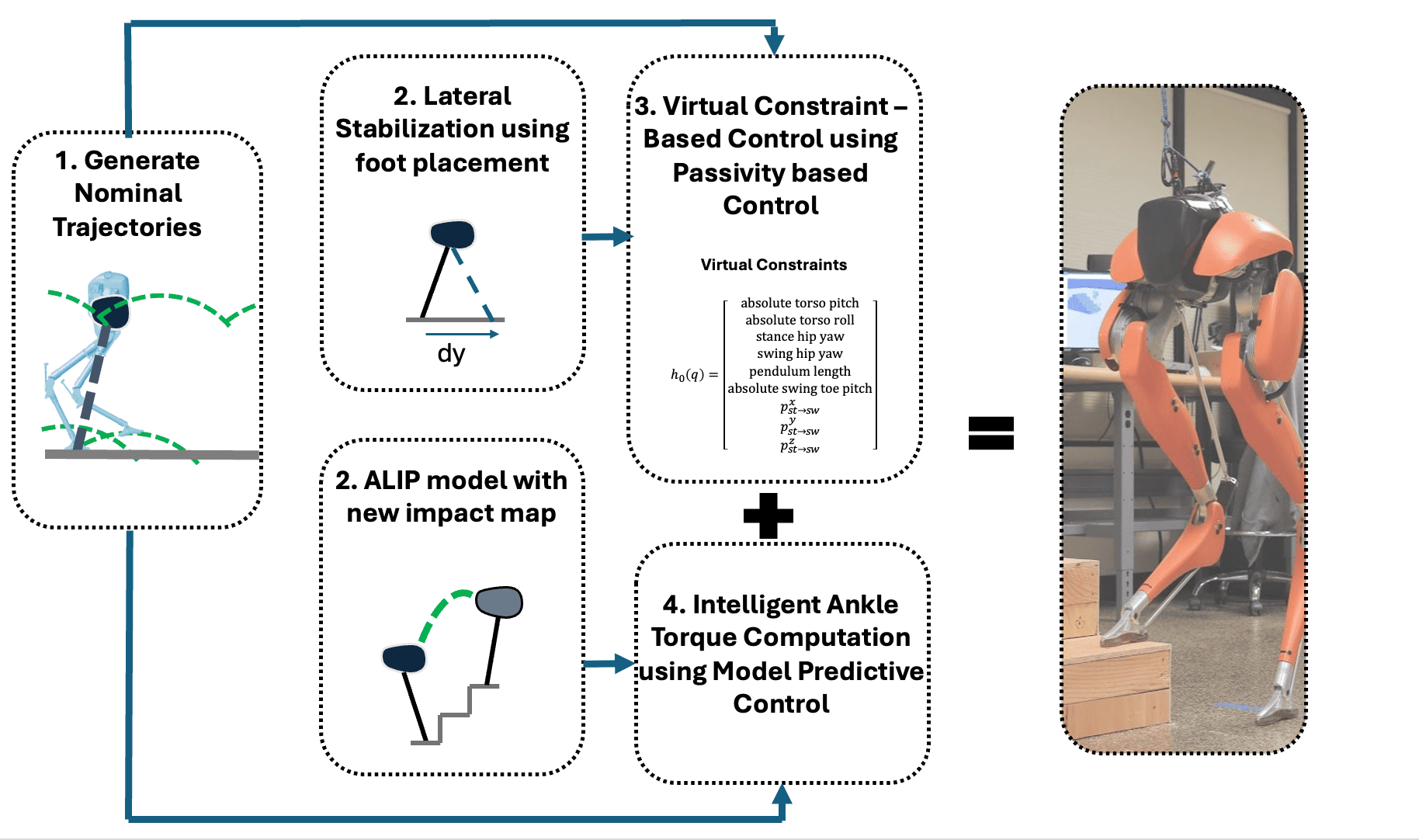}
    \caption{Schematic diagram of the multi-terrain locomotion controller.}
    \label{fig:controller_diagram}
\end{figure}

%%%%%%%%%%%%%%%%%%%%% Control Philosophy %%%%%%%%%%%%%%%%%%%%%%%
\subsection{Review of Controller Design Philosophy}
\label{sec:controlPhilosoph}
Firstly, we review the control philosophy.
% from our previous work \cite{dosunmu2023stair} and \cite{dosunmu2024demonstrating}.
Cassie has 20 degree-of-freedom (DoF) to control. During single support, with one foot grounded and the other free, nine DoF are constrained: four from Cassie's springs (two per leg) and five from the stance foot, leaving 11 DoF to control with ten actuators, making the robot underactuated.

We use Passivity-Based Control (PBC) to regulate nine DoF, utilizing nine of Cassie's actuators to track offline-generated nominal trajectories. The tenth actuator controls one DoF, the stance toe actuator, via a model predictive controller (MPC). The final DoF is managed through a lateral foot placement approach.

\subsection{Nominal Periodic Orbit} \label{sec:nominal_trajectories}
A nominal periodic orbit refers to a predefined, idealized trajectory that a bipedal robot follows during cyclic locomotion tasks, such as walking or stair climbing. We used the Fast Robot Optimization and Simulation Toolkit (FROST) \cite{Hereid2017FROST} to generate nominal trajectories for Cassie. FROST optimizes and simulates the full-body dynamics of bipedal robots, integrating virtual constraints-based feedback controllers and employing a Wolfram Mathematica backend for symbolic generation of multi-domain system dynamics and kinematics. These symbolic expressions are translated into C/C++ code and compiled into *.MEX files for MATLAB, enhancing computational speed for hardware implementation.

FROST models dynamic bipedal walking as a hybrid system with continuous phases and discrete transitions, represented by a Directed Graph. It uses direct collocation for gait optimization, ensuring swift and reliable convergence. We generated multiple nominal trajectories for Cassie, including scenarios like marching in place, walking forward, and transitioning to different inclines. We approximated optimized trajectories using Bézier curves \cite{farin1983algorithms}, essential for MPC computation for ankle torque.

Incorporating Bézier curves into our controller for representing nominal trajectories from FROST allows fine manipulation through control point adjustments. For example, to extend the swing foot's ascent time and avoid an obstacle, we reposition the interior control points of its sagittal plane motion closer to the initial point. This adjustment prolongs the swing foot's sagittal position as it elevates in the z-plane. Bézier curves enable nuanced refinements to nominal trajectories without needing time-intensive offline optimization.

%%%%%%%%%%%%%%%%%%%%%%%% Passivity-Based Control %%%%%%%%%%%%%%%%%%%%%%%%%%%%%
\subsection{Passivity-Based Control}
\label{sec:pbc}

Passivity-Based Control (PBC) is a powerful strategy for controlling nonlinear systems like bipedal robots \cite{sadeghian2017passivity}. It is particularly practical for hardware applications as it does not rely on an accurate system model. This inherent robustness helps protect against sensor imperfections and uncertainties in the robot's kinematic and dynamic properties.

We design a passivity-based controller as follows:
\begin{equation}
    \bar{D}\ddot{y} + (\bar{C} + K_D) \dot{y} + K_P y = 0.
    \label{eq:y_law}
\end{equation}
Here $\bar{D}$ and $\bar{C}$ are related to the system dynamics, while $K_D$ and $K_P$ are user-defined derivative and proportional controller gains, respectively. The output function $y(x)$ is defined by:
\begin{equation}
    y(x) = h_0(q) - h_d(q, p^{x~des}_{sw}, p^{y~des}_{sw},p^{z~des}_{sw},t)
    \label{eq:y}
\end{equation}
where $h_0$ is the collection of virtual constraints and $h_d$ provides the desired trajectories for the virtual constraints. The virtual constraints are defined as follows:
\begin{equation}
    h_0(q) = 
    \begin{bmatrix}
    \text{absolute torso pitch}\\
    \text{absolute torso roll}\\
    \text{stance hip yaw}\\
    \text{swing hip yaw}\\
    \text{pendulum length}\\
    p_{st \to sw}^x\\
    p_{st \to sw}^y\\
    p_{st \to sw}^z\\
    \text{absolute swing toe pitch}
    \end{bmatrix}
    \label{eq:h}
\end{equation}
where the pendulum length describes the vector $r_c$ from the stance foot to the CoM and $p_{st \to sw}$ is the vector emanating from the stance foot and ending at the swing foot.

%%%%%%%%%%%%%%%%%%%%%%%%%%%%% Model Predictive Control %%%%%%%%%%%%%%%%%%%%%%%%%%%%%%%%%%
\subsection{Model Predictive Control using Quadratic Programming}
\label{sec:mpc}

The premise of \textit{Model Predictive Control} (MPC) is to leverage a model of the system to predict its evolution over a defined time horizon, enabling the computation of optimal control inputs to achieve a desired goal state. The MPC framework optimizes ankle torque over a discretized horizon to ensure stability on uneven terrain.

% \subsubsection{Discrete-time Model Formulation}

\begin{figure}
    \centering
    \includegraphics[width=0.15\textwidth]{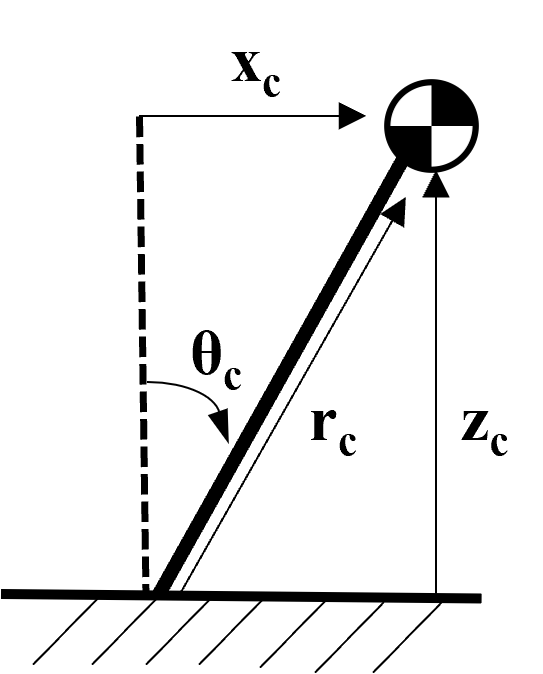}
    \caption{Schematic of an inverted pendulum used to derive a variation of the ALIP model.}
    \label{fig:ARIP}
\end{figure}

We use the below mentioned variation of ALIP for the MPC formulation because it does not assume a constant center-of-mass (CoM) height, which is necessary for the robot to climb stairs.  

Assume an inverted pendulum as shown in Fig.~\ref{fig:ARIP}, where $(x_c, z_c)$ is the Cartesian position of the CoM with respect to the stance foot. The angle of the CoM relative to the stance foot is
\[
    \theta_c = \arctan\left(\frac{x_c}{z_c}\right),
\]
and the pendulum length is
\[
    r_c = \sqrt{x_c^2 + z_c^2}.
\]

Then the ALIP model is given by the following equations:
\begin{equation}
\begin{aligned}
    \dot{\theta}_c &= \frac{L}{m r_c^2(t)} \\
    \dot{L} &= m g r_c \theta_c + \tau,
\end{aligned}
\label{eq:newALIP}
\end{equation}
where $m$ denotes the total mass, $g$ is gravity, and $\tau$ is the torque about the contact point, referred to as stance ankle torque. The ALIP model simplifies bipedal walking dynamics to an inverted pendulum.

This variation of the ALIP model is time-varying because we assume that $r_c(t)$ evolves according to the nominal periodic orbit. For the derivation of this model, refer to \cite{dosunmu2023stair}.

We define the state of our system, represented by \eqref{eq:newALIP}, as \( x(t) = \left[ \theta_c(t) ~~ L(t) \right]^\top \), and discretize it.

We chose to have a higher weight for the impact state than normal state because we noted better controller stability when the robot better tracks the nominal trajectory at impact than when it better tracks the normal state. We define our state error variable as
\begin{equation}\label{eq:x_error}
\begin{split}
& x' = x-x^{des} \\
& x'^{imp} = x^{imp}-x^{imp_{des}}\\
\end{split}
\end{equation}
where $x'$ is the error between current state $x$ and desired state $x^{des}$ during the continuous domain of our hybrid dynamics. $x'^{imp}$ is the error between current state $x^{imp}$ and desired state $x^{imp_{des}}$ just after impact/domain switch of our hybrid dynamics. We define our quadratic program as
\begin{equation}\label{eq:QP}
\begin{split}
& \min_{u_{seq}} \sum_{k=1}^{N} \Bigg[u^T_{k} H u_{k} + ({x'}_k^\top Q {x'}_k + {x'}_k^{imp\top} Q_{imp} {x'}_k^{imp})\Bigg] \\
&\text{subject to} \\
& x_{k+1} = A_k x_k + b_k u_k \\
    & u_{min} < u_k < u_{max}.
\end{split}
\end{equation}
where $k ( 0\leq k \leq N) $ denotes the time index with N the prediction horizon length, $A_k$ and $b_k$ are derived by discretizations of the system dynamics, $u_k \in \mathbb{R}$ is the ankle torque $\tau$, $H$ is weight matrix for the control variable $u$, and $Q$ is the weight matrix for $x'$ and $Q_{imp}$ the weight matrix for $x'^{imp}$.
We want to minimize the torque sequence \( u_k^{seq} \) with stringent bounds \( u_{min} \) and \( u_{max} \) ensuring physical feasibility.

We used CasADi for our implementation of MPC in C language. CasADi \cite{Andersson2019} is an open-source tool for nonlinear optimization and algorithmic differentiation. After writing the QP formulation in C language, we converted it to dynamically linked library (DLL) and solved the QP using the QRQP solver with SQP method. We chose the SQP method with QRQP because it delivers more accuracy in dual variables when it converges.

Our MPC selects the trajectory it needs to follow depending on the terrain. Using FROST, we generated trajectories for flat ground and $4^{\circ}$, $8^{\circ}$, $15^{\circ}$ and $20^{\circ}$ inclined plane and stairs with the step height of 20 cm and step length of 25 cm. Our decision variables include the $\theta_c$ and $L_y$ and $\tau$. The moving horizon is one foot step ahead with the trajectory discretized at 50 intervals. We used a multiple-shooting-based numerical optimization for our problem. This MPC formulation enables Cassie to adapt ankle torque in real time for stair climbing (Section \ref{sec:results_stairs}).

%%%%%%%%%%%%%%%%%%%%%%%%%%%% Lateral Stablization %%%%%%%%%%%%%%%%%
\subsection{Lateral Stabilization of the Robot}
\label{sec:lateralStabilization}

We stabilized Cassie's lateral motion using an angular momentum-based foot placement strategy adapted from \cite{gong2021}. We can use this foot placement strategy in lateral placement since stairs are generally wide. Moreover Cassie's ankles does not have actuation in the lateral direction. Unlike its predecessor, the new ALIP lacks a closed-form solution, necessitating a numerical approach. To ensure real-time feasibility on hardware, we implemented a look-up table, enabling execution times under 5 $\mu$s—critical to meet Cassie's 2 kHz control-update frequency.

The numerical strategy for computing lateral foot placement is:
\begin{enumerate}
    \item Use Euler's method for numerical integration to estimate angular momentum before impact for two nearby lateral foot positions, $y_{\text{st}\rightarrow\text{sw}}^1$ and $y_{\text{st}\rightarrow\text{sw}}^2$, denoted as $L_1$ and $L_2$.
    \item Define a desired angular momentum $L_{\text{des}}$ approximated along the line between $L_1$ and $L_2$.
    \item Employ 2D or 3D linear interpolation to find $y_{\text{des}}$, the lateral foot placement corresponding to $L_\text{des}$.
\end{enumerate}

The full derivation of this approach presented in \cite{dosunmu2024demonstrating}.

% CasADi
% \input{Sections/CasADi/casadi}

% UDP
% \input{Sections/ImplementingUDP/ImplementingUDP}

%ALIP 
% \input{Sections/ALIP/ALIP}
% MPC Impact
\section{Developing an Impact Map by Linearizing over a Nominal Trajectory}
\label{sec:impactReducedOrder}

To mitigate torque instabilities on compliant terrain, we derive a new impact map based on linearized nominal trajectories. The impact map of the ALIP model for the MPC is based on linearization about a nominal trajectory. We developed this impact map to reduce the spikes in ankle torque computed by the MPC controller and to add robustness to the controller.

We  develop an impact map based off a linearization about the nominal trajectory for center of mass (CoM) angle, $\bar{\theta}_c^+$, and angular momentum. We start with CoM angle.

\subsection{CoM Angle}
\begin{figure}
    \centering
    \includegraphics[width=.40\textwidth]{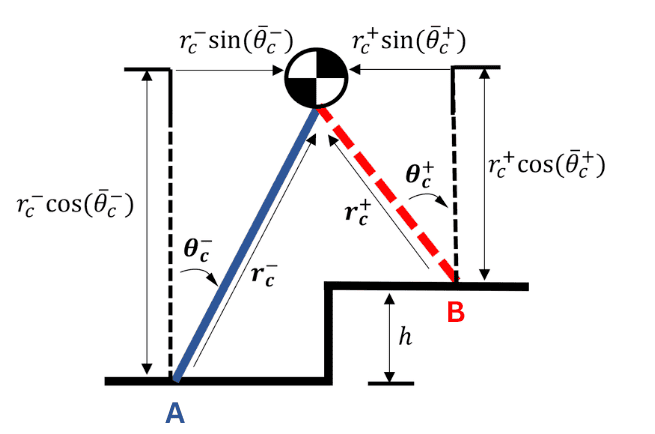}
    \caption{Schematic depicting geometry variables for computing impact map based off of a lineariziation of a nominal trajectory.}
    \label{fig:impactForLinearizing}
\end{figure}

Figure \ref{fig:impactForLinearizing} depicts the relevant variables for computing the impact map. Note that $\bar{\theta}_c^+$ refers to the CoM angle after impact and $\bar{\theta}_c^-$ refers to CoM angle before impact. 

From geometry, we have
\begin{equation}
    r_c^- \cos(\bar{\theta}_c^-) = r_c^+\cos(\bar{\theta}_c^+) + h.
    \label{eq:rcminuscos}
\end{equation}

While we wish to perfectly follow the nominal trajectory $\bar{\theta}_c^+$, in reality, we do not. Instead, we have an error $\delta \theta_c$. Thus, \eqref{eq:rcminuscos} becomes
\begin{equation}
    r_c^- \cos(\bar{\theta}_c^- + \delta \theta_c^-) = r_c^+\cos(\bar{\theta}_c^+ + \delta \theta_c^+) + h.
    \label{eq:rcminuscosFinal}
\end{equation}

Applying linearization on $\cos(\bar{\theta}_c^- + \delta \theta_c^-)$ and $\cos(\bar{\theta}_c^+ + \delta \theta_c^+)$ on \eqref{eq:rcminuscosFinal} and simplifying yields

\small
\begin{align}
    r_c^- \cos(\bar{\theta}_c^-) - r_c^- \sin(\bar{\theta}_c^-)\delta \theta_c^-  &= r_c^+ \cos(\bar{\theta}_c^+) - r_c^+  \sin(\bar{\theta}_c^+)\delta \theta_c^+ + h
    \label{eq:rcminuscossubbed}
\end{align}
\normalsize

Plugging in \eqref{eq:rcminuscos} into \eqref{eq:rcminuscossubbed} and simplifying yields
\begin{align}
    - r_c^- \sin(\bar{\theta}_c^-) \delta \theta_c^- &=  - r_c^+ \sin(\bar{\theta}_c^+) \delta \theta_c^+ .
\end{align}

Solving for the error $\delta \theta_c^+$ yields
\begin{equation}
    \delta \theta_c^+ = \frac{r_c^- \sin(\bar{\theta}_c^-)}{r_c^+ \sin(\bar{\theta}_c^+)} \delta \theta_c^-,
    \label{eq:errorThetaForTaylor}
\end{equation}
where $r_c^- \sin(\bar{\theta}_c^+)$ is a constant given by the nominal trajectory.

The actual value of $\theta_c^+$ is given by, 
\begin{equation}
    \theta_c^+ = \delta \theta_c^+ + \bar{\theta}_c^+.
    \label{eq:thetaForTaylor}
\end{equation}

Plugging in \eqref{eq:errorThetaForTaylor} into \eqref{eq:thetaForTaylor} yields
\begin{equation}
    \theta_c^+ = \frac{r_c^- \sin(\bar{\theta}_c^-)}{r_c^+ \sin(\bar{\theta}_c^+)} \delta \theta_c^- + \bar{\theta}_c^+.
    \label{eq:thetacPlusImpact}
\end{equation}

\subsection{Angular Momentum}
Generally, angular momentum about the contact point is invariant during impact but since our ankle is actuated the angular momentum about the contact point is not constant. So we need compute the impact map for angular momentum by linearizing about its nominal trajectory. We start by using the impact map derived for angular momentum in \cite{dosunmu2024demonstrating}:

\small
\begin{equation*}
    x^+ = 
    \begin{bmatrix}
        \theta_c^+ \\
        L^+
    \end{bmatrix}
\end{equation*}
\begin{equation}
    =
    \begin{bmatrix}
        \arccos \Bigg( \frac{r_c^- \cos \theta_c^- - P_\text{st $\rightarrow$ sw}^z}{r_c^+} \Bigg) \\
        \begin{matrix}
            L_B^- + m \Bigg[P_\text{st $\rightarrow$ sw}^z \Big(r_c^- \cos (\theta_c^-) \dot{\theta}_c^- + \dot{r}_c^- \sin(\theta_c^-) \Big)\\
           - P_\text{st $\rightarrow$ sw}^x \Big(-r_c^- \sin (\theta_c^-) \dot{\theta}_c^- + \dot{r}_c^- \cos(\theta_c^-) \Big) \Bigg] 
        \end{matrix}
    \end{bmatrix}. \label{eq:LAplus}
\end{equation}
\normalsize

From \eqref{eq:LAplus}, we know that if we were perfectly following the nominal trajectory, the angular momentum after impact would be
\begin{equation}
    \begin{aligned}
        \bar{L}_A^+ = \bar{L}_B^- + m \Bigg[P_\text{st $\rightarrow$ sw}^z \Big(r_c^- \cos (\bar{\theta}_c^-) \dot{\bar{\theta}}_c^- + \dot{r}_c^- \sin(\bar{\theta}_c^-) \Big) - \\
        P_\text{st $\rightarrow$ sw}^x \Big(-r_c^- \sin (\bar{\theta}_c^-) \dot{\bar{\theta}}_c^- + \dot{r}_c^- \cos(\bar{\theta}_c^-) \Big) \Bigg] .
    \end{aligned}
    \label{eq:LAplusLinearization}
\end{equation}

In reality, we have error $\delta L$ and $\delta \theta_c$. Thus, \eqref{eq:LAplusLinearization} becomes
\begin{equation}
\begin{aligned}
    (\bar{L}_A^+ + \delta L^+) &= (\bar{L}_B^- + \delta L^-) \\
    & \quad + m \Bigg[P_{\text{st} \rightarrow \text{sw}}^z \Big(r_c^- \cos (\bar{\theta}_c^- +\delta \theta_c^-) (\dot{\bar{\theta}}_c^- + \delta \dot{\theta}_c) \\
    & \quad + \dot{r}_c^- \sin(\bar{\theta}_c^- + \delta \theta_c^-) \Big) \\
    & \quad - P_{\text{st} \rightarrow \text{sw}}^x \Big(-r_c^- \sin (\bar{\theta}_c^- + \delta \theta_c^-) (\dot{\bar{\theta}}_c^- + \delta \bar{\theta}_c^-) \\
    & \quad + \dot{r}_c^- \cos(\bar{\theta}_c^- + \delta \theta_c^-) \Big) \Bigg].
\end{aligned}
\label{eq:LAplusLinearizationError}
\end{equation}

We linearize \eqref{eq:LAplusLinearizationError} using similar calculations as we did for the CoM angle. We arrive at the following:

\begin{equation}
\begin{aligned}
    L^+ &= \bar{L}_B^- + \delta L^- - \bar{L}_A^+ \\
    &\quad + m \Bigg[P_{\text{st} \rightarrow \text{sw}}^z \Big(r_c^- \Big[ \cos (\bar{\theta}_c^-) - \delta \theta_c^- \sin(\bar{\theta}_c^-) \Big] (\dot{\bar{\theta}}_c^-) \\
    &\quad + \dot{r}_c^- \Big[ \sin(\bar{\theta}_c^-) + \delta \theta_c^- \cos(\bar{\theta}_c^-) \Big] \Big) \\
    &\quad - P_{\text{st} \rightarrow \text{sw}}^x \Big(-r_c^- \Big[ \sin (\bar{\theta}_c^-) + \delta \theta_c^- \cos(\bar{\theta}_c^-) \Big] (\dot{\bar{\theta}}_c^-) \\
    &\quad + \dot{r}_c^- \Big[ \cos(\bar{\theta}_c^-) - \delta \theta_c^- \sin(\bar{\theta}_c^-) \Big] \Big) \Bigg] \\
    &\quad + \bar{L}^+.
\end{aligned}
\label{eq:LAplusImpact}
\end{equation}

Equation \eqref{eq:thetacPlusImpact} and \eqref{eq:LAplusImpact} form our impact map for the reduced-order model. This linearized impact map directly addresses torque spikes observed in prior work \cite{dosunmu2024demonstrating}, enabling stable locomotion on gravel and sand as shown in Section \ref{sec:results_sand}.

% Results 
\section{Hardware Results and Discussion} \label{section:Hardware}
Having derived the impact map and controller structure, we now present hardware validation across diverse terrains using 20 DOF Cassie bipedal robot.
% This section discusses the practical implementation of the enhanced controller on the physical 20 DOF Cassie bipedal robot hardware.
% While the gait SimMechanics simulations in \cite{dosunmu2023stair} and preliminary hardware implementations in \cite{dosunmu2024demonstrating} were promising, a key challenge for bipedal robots is navigating unstructured environments. 
We strategically incorporate experiments that move away from the safe, structured lab environment into the uncontrolled and unstructured outdoor environment. The new experiments aim to evaluate Cassie's stability and adaptability on various natural terrains, extending its operational capabilities beyond the controlled indoor environments. This evaluation provides a more rigorous test of the controller's robustness and practical utility in real-world settings.

\subsection{Cassie walks on sand, gravel, and rocks} \label{sec:results_sand}
\begin{figure}
    \centering
    \includegraphics[width=.3\textwidth]{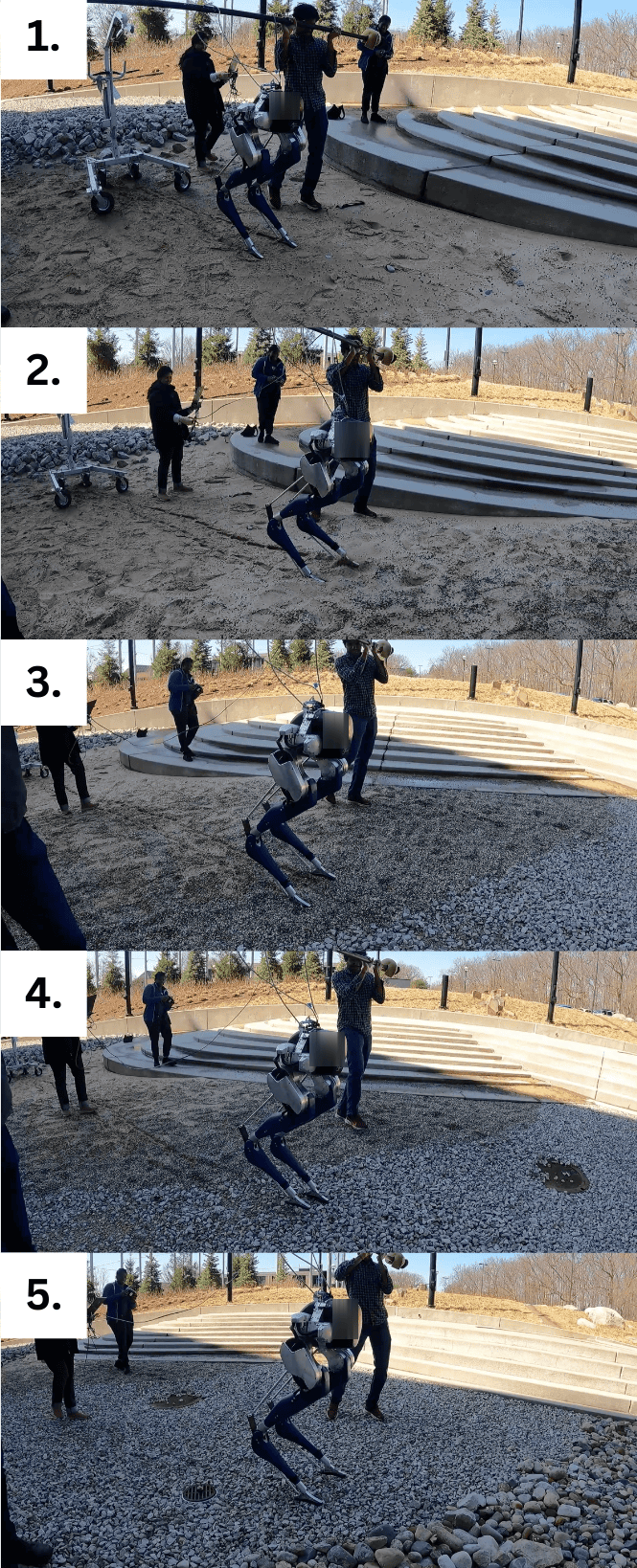}
    \caption{A series of images depicting hardware results of Cassie bipedal robot walking through the Robot Playground, where the robot walked continuously through sand, gravel, and rocks. The original video can be found in \cite{DynamicLegLocomotion_sandgrassgravel}}
    \label{fig:cassie_sand_stone}
\end{figure}

Hard surfaces like concrete and treadmills provide ideal conditions for testing controllers. However, a key benchmark for controller performance is the ability to transition from safe indoor environments to unstructured outdoor settings successfully. To evaluate Cassie equipped with the enhanced controller discussed in this paper, we conducted rigorous tests to assess its stability on various loose terrains. We navigated Cassie through the Robot Playground, where the robot walked continuously through sand, gravel, and rocks as shown in Figure \ref{fig:cassie_sand_stone}. The video for the experiment can be found in \cite{DynamicLegLocomotion_sandgrassgravel}. Figure \ref{fig:cassie_sand_stone} shows the Cassie transitioning from sand to gravel to finally small stones. Due to the line foot of the Cassie robot, we could not test our controller on larger rocks. 

The development of the new impact map developed in Section \ref{sec:impactReducedOrder} played a key role in the success of this experiment by addressing the issue of spiking ankle torque values which would not be ideal in uneven and compliant terrain. This new impact map linearizes about the nominal trajectory, effectively reducing the spikes that occurred with the previous impact map in \cite{dosunmu2024demonstrating} calculated using the dynamics of a reduced-order model. While large spikes in ankle torque were manageable on hard surfaces, such as concrete and treadmills, they posed significant challenges on loose terrains like sand, gravel, and rocks. On these less stable surfaces, an accurate impact map is critical to maintaining balance and stability, ensuring that Cassie could walk effectively and safely through the varied outdoor environments tested in this study.

%%%%%%%%%%%%%%%%%%%%%%%%%%%%% Steep, wet grassy slope %%%%%%%%%%%%%%%%%%
\subsection{Cassie marches up a steep, wet grassy slope} \label{sec:results_slope}
\begin{figure}
    \centering
    \includegraphics[width=.30\textwidth]{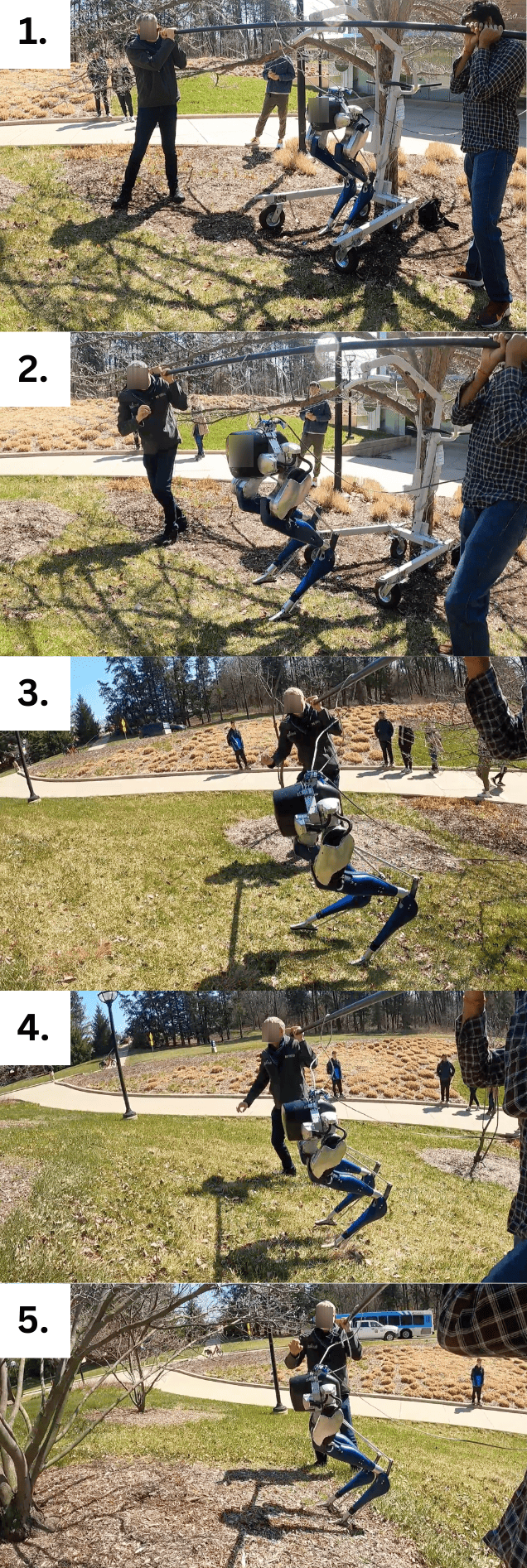}
    \caption{A series of images depicting hardware results of Cassie bipedal robot walking on grassy slope with an estimated average gradient of approximately 22 degrees from a side view perspective. The original video can be found in \cite{DynamicLegLocomotion_grassyslope}}
    \label{fig:cassie_slope}
\end{figure}

In the next experiment, we evaluated Cassie's performance on a challenging outdoor slope to further validate the enhanced controller. The video for this experiment can be found in \cite{DynamicLegLocomotion_grassyslope}. This test was conducted on a steep, wet grassy incline as shown in Figure \ref{fig:cassie_slope}, with an estimated average gradient of approximately 22 degrees. Figure \ref{fig:cassie_slope} shows the progression of walking on the uneven steep slope. The lateral stabilization strategy (Section \ref{sec:lateralStabilization}) allowed Cassie to maintain balance on the 22-degree wet slope. Without this adaptation, angular momentum errors would accumulate, leading to falls on such steep, slippery terrain. While Cassie had previously ascended this slope using earlier controllers \cite{gibson2022terrain}, successfully navigating it with our controller discussed in this paper represented a significant milestone before progressing to stair-climbing tests. The slope's uneven and rain-slicked surface provided a rigorous and realistic environment, essential for demonstrating the controller's robustness. 

%%%%%%%%%%%%%%%%% Shallow stairs %%%%%%%%%%%%%%%%%%%%%%%%%
\subsection{Cassie blindly climbs short stairs} \label{sec:results_stairs}
\begin{figure}
    \centering
    \includegraphics[width=.20\textwidth]{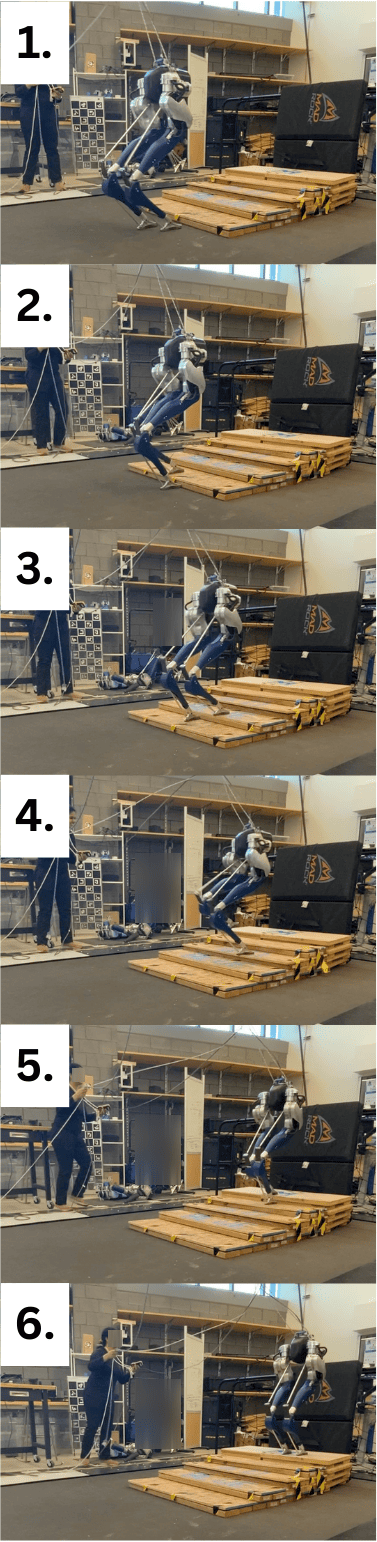}
    \caption{A series of images depicting hardware results of Cassie bipedal robot walking on shallow stairs from a back view perspective. Each step is 2.5 inches in height, with a depth of 12 inches. The original video can be found in \cite{DynamicLegLocomotion_shallowsteps}}
    \label{fig:cassie_stair}
\end{figure}

In the final experiment, we assessed Cassie's ability to navigate a staircase in a lab environment as shown in Figure \ref{fig:cassie_stair} which depicts the progression of stepping on the stairs. The video of the experiment can be found in \cite{DynamicLegLocomotion_shallowsteps}. The nominal trajectories generated using FROST (Section \ref{sec:nominal_trajectories}) ensure swing-foot clearance over stair edges. During the experiment, Cassie ascended the stairs without the aid of visual or pre-programmed foothold data, resulting in sub-optimal foot placement. Despite this, Cassie successfully climbed to the top of the 5-step staircase. Each step with 2.5 inches tall, with a depth of 12 inches.

% Conclusions
\section{Discussion and Conclusions}
\label{sec:conclusions}

In this paper, we present a robust controller for an underactuated bipedal robot capable of blindly navigating on rigid uneven, compliant, and constrained terrains. We present a holistic description of the entire control structure, the derivation of a new impact model for the reduced-order model, and hardware demonstration of the improved controller over sand, gravel, rocks, and a steep, uneven wet grassy slope and stairs. Our holistic controller structure enabled seamless transitions between terrains, while the impact map eliminated torque spikes critical for outdoor navigation.

While the results presented in this paper are promising, Cassie's ability to navigate its environment blindly presents some limitations. For instance, during the stair-climbing experiment, Cassie successfully ascended five steps; however, this relied on the robot operator precisely timing the transition from walking forward to climbing stairs. Despite the operator's timing being off, resulting in suboptimal foot placement, the controller's robustness enabled successful completion of the task. Nevertheless, for Cassie to climb more stairs consecutively and handle stairs of varying heights and depths, the controller requires perception information for accurate foot placement. In future work, we will integrate a perception system, such as the one described in \cite{Huang2023}, enabling Cassie to autonomously navigate stairs.

\bibliographystyle{IEEEtran}
\bibliography{references.bib}
% \bibitem{b1} G. Eason, B. Noble, and I. N. Sneddon, ``On certain integrals of Lipschitz-Hankel type involving products of Bessel functions,'' Phil. Trans. Roy. Soc. London, vol. A247, pp. 529--551, April 1955.
% \bibitem{b2} J. Clerk Maxwell, A Treatise on Electricity and Magnetism, 3rd ed., vol. 2. Oxford: Clarendon, 1892, pp.68--73.
% \bibitem{b3} I. S. Jacobs and C. P. Bean, ``Fine particles, thin films and exchange anisotropy,'' in Magnetism, vol. III, G. T. Rado and H. Suhl, Eds. New York: Academic, 1963, pp. 271--350.
% \bibitem{b4} K. Elissa, ``Title of paper if known,'' unpublished.
% \bibitem{b5} R. Nicole, ``Title of paper with only first word capitalized,'' J. Name Stand. Abbrev., in press.
% \bibitem{b6} Y. Yorozu, M. Hirano, K. Oka, and Y. Tagawa, ``Electron spectroscopy studies on magneto-optical media and plastic substrate interface,'' IEEE Transl. J. Magn. Japan, vol. 2, pp. 740--741, August 1987 [Digests 9th Annual Conf. Magnetics Japan, p. 301, 1982].
% \bibitem{b7} M. Young, The Technical Writer's Handbook. Mill Valley, CA: University Science, 1989.
% \end{thebibliography}
% \vspace{12pt}
% \color{red}
% IEEE conference templates contain guidance text for composing and formatting conference papers. Please ensure that all template text is removed from your conference paper prior to submission to the conference. Failure to remove the template text from your paper may result in your paper not being published.

\end{document}